# Quality Detection of Stored Potatoes via Transfer Learning: A CNN and Vision Transformer Approach

## Authors

Shrikant Kapse*, Priyankkumar Dhrangdhariya, Priya Kedia, Manasi Patwardhan, Shankar Kausley, Soumyadipta Maiti, Beena Rai, Shirish Karande

*TCS Research, Phase 3, Hinjawadi Rajiv Gandhi Infotech Park, Hinjawadi, Pune, Pimpri-Chinchwad, Maharashtra, India-411057*

*Corresponding author: TCS Research, Phase 3, Hinjawadi Rajiv Gandhi Infotech Park, Hinjawadi, Pune, Pimpri-Chinchwad, Maharashtra, India- 411057

E-mail address: shrikant.kapse@tcs.com

**Abstract**

Image-based deep learning provides a non-invasive, scalable solution for monitoring potato quality during storage, addressing key challenges such as sprout detection, weight loss estimation, and shelf-life prediction. In this study, images and corresponding weight data were collected over a 200-day period under controlled temperature and humidity conditions. Leveraging powerful pre-trained architectures of ResNet, VGG, DenseNet, and Vision Transformer (ViT), we designed two specialized models: (1) a high-precision binary classifier for sprout detection, and (2) an advanced multi-class predictor to estimate weight loss and forecast remaining shelf-life with remarkable accuracy. DenseNet achieved exceptional performance, with 98.03% accuracy in sprout detection. Shelf-life prediction models performed best with coarse class divisions (2–5 classes), achieving over 89.83% accuracy, while accuracy declined for finer divisions (6–8 classes) due to subtle visual differences and limited data per class. These findings demonstrate the feasibility of integrating image-based models into automated sorting and inventory systems, enabling early identification of sprouted potatoes and dynamic categorization based on storage stage. Practical implications include improved inventory management, differential pricing strategies, and reduced food waste across supply chains. While predicting exact shelf-life intervals remains challenging, focusing on broader class divisions ensures robust performance. Future research should aim to develop generalized models trained on diverse potato varieties and storage conditions to enhance adaptability and scalability. Overall, this approach offers a cost-effective, non-destructive method for quality assessment, supporting efficiency and sustainability in potato storage and distribution.

## 1. Introduction

Potato is the world's third most cultivated food crop after rice and wheat, serving as a staple for billions and playing a vital role in global food security. However, during post-harvest storage and distribution, potatoes undergo significant physiological and biochemical changes influenced by environmental conditions such as temperature, humidity, surrounding gas concentration and storage duration. These changes manifest in both internal and external quality parameters, including sugar content, weight loss, pH, total soluble solids, and visible surface alterations like sprouting, wrinkling, fungal growth, and skin darkening. Among these, weight loss and sprouting are critical indicators of quality degradation. Loss of weight, mainly caused by transpiration and respiration, results in noticeable wrinkling and a decrease in marketable mass. Potatoes that lose more than 10% of their weight are often deemed unfit for sale (Pinhero et al., 2009). Sprouting, driven by hormonal changes involving gibberellins, cytokinins, and abscisic acid (ABA), not only accelerates weight loss but also diminishes consumer acceptability (Sonnewald & Sonnewald, 2014). The sprout's epidermis is significantly more permeable to water than the surface, further increasing transpiration rates.

Given that fresh potatoes are available for only 3-4 months annually, stored potatoes dominate the market for the rest of the year. These stored tubers vary in age and quality, necessitating effective methods for quality-based segregation to support differential pricing, inventory management, and timely sale or purchase decisions. Traditional manual inspection methods are labor-intensive, subjective, and inefficient for large-scale operations. Recent advances in image-based deep learning have shown promise in automating postharvest quality assessment. Unlike traditional machine learning approaches that rely on handcrafted features and controlled imaging conditions, deep learning, particularly convolutional neural networks (CNNs) and Vision Transformers (ViTs) can automatically learn complex visual patterns such as sprout morphology, surface wrinkles, and discoloration (Wang et al., 2025 & Gülmez et al., 2025). These models have been successfully applied in grading (ripening stage) and defect detection of various crops, including dates, tomatoes, bananas, and potatoes (Piedad et al., 2018; Surya Prabha & Satheesh Kumar, 2015; Nasiri et al., 2019). For instance, CNN-based systems have achieved over 86% accuracy in potato grading using depth imaging, while YOLO-based models have been used to detect external defects like greening and rot in real time (Su et al., 2020; Wu et al., 2025).

Despite these advancements, there is a notable gap in the literature regarding image-based prediction of potato shelf-life and maturity. While some studies have explored moisture content estimation, disease detection, and defect classification, few have addressed the integration of visual cues with weight loss and sprouting data to estimate remaining shelf-life (Kamilaris & Prenafeta-Boldú, 2018; Le & Lin, 2019). Previous research has applied various machine learning and image-processing techniques for potato quality assessment using visual features. Moisture content estimation based on color and texture employed multi-layer perceptron (MLP) models (Ebrahimi et al., 2012). Aspect ratio was analyzed using marker-controlled watershed segmentation (Si et al., 2017), while external defects like rotting and greening were detected using PCA-based feature selection combined with nearest neighbor and genetic algorithms (Dacal-Nieto et al., 2009). Variety identification using color, texture, and shape has also been implemented with PCA, discriminant analysis, and ANN models (Azizi et al., 2016). Wei et al. present a cross-modal detection strategy that integrates machine vision with VIS/NIR spectroscopy to achieve real-time, intelligent classification of both external and internal potato defects (Wei et al., 2024). Imaging spectroscopy, combined with machine learning, has proven to be a powerful non-destructive approach for characterizing and predicting quality attributes of tuber crops (Su and Xue, 2021). Recent integration of depth cameras with convolutional neural networks has demonstrated high accuracy in size and appearance classification, highlighting the potential of 3D imaging for postharvest quality assessment (Su et al., 2020). Automated vision systems have shown strong potential for nondestructive potato quality inspection by leveraging color, size, texture, and firmness features extracted from digital images (Ibrahim et al., 2020). Recent advances in imaging spectroscopy, machine vision, depth imaging, and automated vision systems integrated with machine learning and AI offer rapid, nondestructive solutions for potato quality assessment, enabling accurate grading based on physical, chemical, and surface attributes (Danielak et al., 2023). Explainable AI combined with transfer learning in deep learning frameworks enables accurate and interpretable potato leaf disease classification, achieving over 97% accuracy while enhancing transparency for practical agricultural applications (Alhammad et al., 2025). Vision-based deep learning techniques enable automation, accurate classification of industrial paint and coating defects, reducing reliance on inefficient manual inspection (Dhrangdhariya et al., 2025). These studies demonstrate the potential of image-based approaches for non-destructive potato quality evaluation.

This study aims to bridge that gap by developing a comprehensive deep learning framework for potato quality detection using image data. Specifically, the objectives are to:

1. Classify the presence of sprouts in potato images.

2. Predict the percentage of weight loss and estimate the remaining shelf-life using multi-class classification.

The study also examines how varying the number of output classes on the accuracy of shelf-life prediction, with the goal of identifying an optimal classification granularity for practical deployment in supply chain management.

## 2. Materials and methods

### 2.1 Data collection

Potatoes belonging to the 'FC3' cultivar were obtained from a local farm in Pune, India. After being washed to remove soil and debris, they were cured for two weeks. Following the curing, 18 potatoes were distributed into trays, with six potatoes placed in each tray. The storage condition was maintained at $21 \pm 2°C$ temperature and $70 \pm 10\%$ relative humidity (RH).

Potato images were captured at intervals ranging from 1 to 5 days, depending on the extent of visible changes, using a Nikon Coolpix P100 digital camera mounted on a fixed stand to ensure consistent imaging distance and angle. Each image captured a tray containing six potatoes, which were subsequently cropped manually using the Snipping Tool to isolate individual potato images (as illustrated in figure 1). The resulting cropped images measured approximately 250 × 250 pixels at 96 dpi. Simultaneously, the weight of tray was recorded at the same intervals using a METTLER TOLEDO PB3002-S precision balance over a period of 200 days. The cumulative weight loss percentage for each potato at any given time point was calculated using the following Equation (1):

$$Cumulative\ weight\ loss\ \% = \frac{W_0 - W_t}{W_0} \times 100 \tag{1}$$

Where,

$W_0$ is weight of potato at start of storage

$W_t$ is weight of potato at any time t

The shelf-life was defined as the number of days required for a potato to reach 10% cumulative weight loss. The remaining shelf-life for each potato at any given time was computed using the following Equation (2):

$$Remaining\ shelf\ life\ (days) = Shelf\ life\ (days) - Current\ day \qquad (2)$$

After collecting the image and weight data at regular intervals, each image was labeled with the appropriate output class to enable supervised training of the corresponding classification models

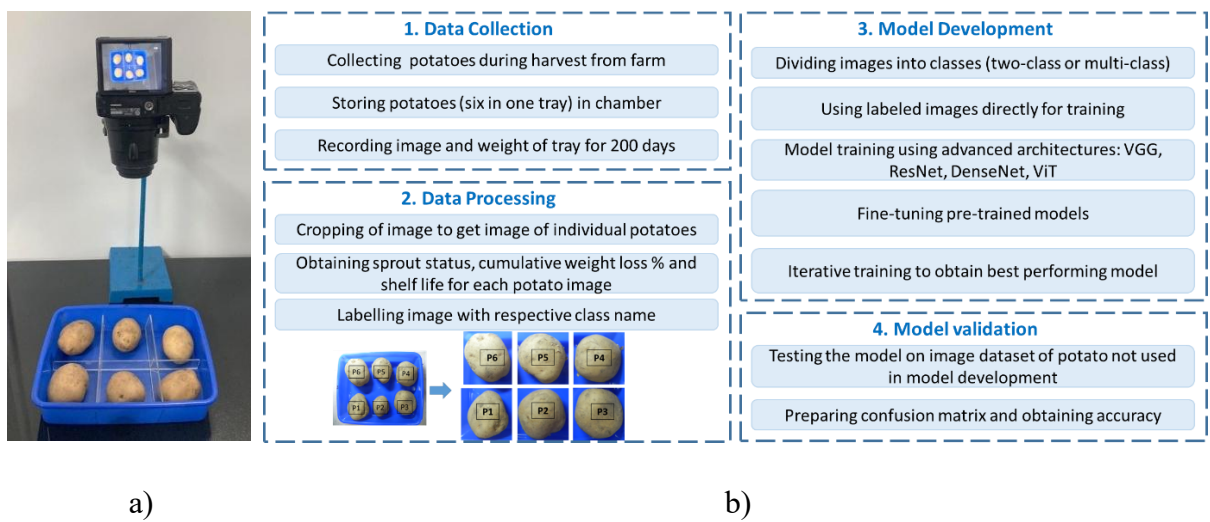

Figure 1: a) Experimental set up for data collection b) The overall architecture of development of image-based models for potatoes

## 2.2 Sprout Classification

The sprout detection model was designed as a binary classification with two classes: 'Sprouts' and 'Non-Sprouts'. The dataset consisted of images of individual potatoes. To ensure robust model evaluation, 80% of the images were allocated for training, while the remaining 20% were reserved for testing. Each image was manually annotated through visual inspection to indicate the presence or absence of sprouts. This process resulted in a labeled dataset comprising 255 training images and 51 test images. Model evaluation was performed using a 5-fold cross-validation approach combined with the GridSearchCV strategy during finetuning of CNN and ViT based models.

## 2.3 Multiclass Shelf-life prediction

Weight loss due to transpiration is a key factor affecting potato quality (shelf-life) during storage. Predicting cumulative weight loss percentage and the corresponding remaining shelf-life from potato images is a complex task, primarily because the visual changes in potatoes occur gradually and are often subtle in the early stages of storage. Initially, potatoes may show little to no visible change, followed by the emergence of small sprouts. As storage time increases, additional sprouts appear, existing ones grow larger, and surface wrinkles begin to form indicating advanced aging. Unlike fruits such as bananas or tomatoes, where age-related changes are often reflected in simple two-dimensional features like color or size, potatoes exhibit more complex three-dimensional visual cues such as sprouting and wrinkling. These features make it challenging to estimate weight loss and shelf-life using traditional image-based methods.

To address this, multi-class classification models were developed to categorize potatoes based on their cumulative weight loss percentage, which correlates with remaining shelf-life. Multiple class configurations (ranging from 2 to 8 classes) were explored, as detailed in Table 1 and sample images of each class as shown in figure 2, to determine the optimal number of classes that balance prediction accuracy with practical usability.

To develop shelf-life prediction models, day-wise images and weight data of potatoes were annotated with the corresponding cumulative weight loss percentage and remaining shelf-life, calculated using Equations (1) and (2). Shelf-life prediction was framed as a multi-class classification problem, with models trained for 2 to 8 class configurations. In each case, the classes represented specific ranges of weight loss percentages. The final class in every configuration included all samples with weight loss greater than 10%, while the remaining classes were created by dividing the 0–10% range equally based on the number of classes minus one (as detailed in Table 1).

For each classification configuration, the dataset was annotated accordingly, and the images were organized into separate directories corresponding to their assigned class labels, with each directory representing a distinct class utilized for training the classification models. The dataset comprised images of six individual potatoes. Model evaluation was performed using a 5-fold cross-validation approach combined with the GridSearchCV strategy. In this setup, 255 images were used for training, while 51 images were reserved for testing.

**Table 1:** Class distribution across multiclass shelf-life prediction models for potatoes

| No. of Classes | Class | Cumulative Weight Loss (%) | Remaining Shelf Life (Days) |
|---|---|---|---|
| 2 | Class I | 0 – 10 | 1 – 121 |
|   | Class II | >10 | 0 |
| 3 | Class I | 0 – 5 | 37 – 121 |
|   | Class II | 5 – 10 | 1 – 36 |
|   | Class III | >10 | 0 |
| 4 | Class I | 0 – 3.3 | 52 – 121 |
|   | Class II | 3.3 – 6.6 | 20 – 51 |
|   | Class III | 6.6 – 9.9 | 1 – 19 |
|   | Class IV | >10 | 0 |
| 5 | Class I | 0 – 2.5 | 62 – 121 |
|   | Class II | 2.5 – 5 | 38 – 61 |
|   | Class III | 5 – 7.5 | 14 – 37 |
|   | Class IV | 7.5 – 10 | 1 – 13 |
|   | Class V | >10 | 0 |
| 6 | Class I | 0 – 2 | 74 – 121 |
|   | Class II | 2 – 4 | 48 – 73 |
|   | Class III | 4 – 6 | 28 – 47 |
|   | Class IV | 6 – 8 | 12 – 27 |
|   | Class V | 8 – 10 | 1 – 11 |
|   | Class VI | >10 | 0 |
| 7 | Class I | 0 – 1.66 | 82 – 121 |
|   | Class II | 1.66 – 3.32 | 52 – 81 |
|   | Class III | 3.32 – 4.98 | 38 – 51 |
|   | Class IV | 4.98 – 6.64 | 21 – 37 |
|   | Class V | 6.64 – 8.3 | 11 – 20 |
|   | Class VI | 8.3 – 9.96 | 1 – 10 |
|   | Class VII | >10 | 0 |
| 8 | Class I | 0 – 1.43 | 89 – 121 |
|   | Class II | 1.43 – 2.86 | 56 – 88 |
|   | Class III | 2.86 – 4.29 | 46 – 55 |
|   | Class IV | 4.29 – 5.72 | 29 – 45 |
|   | Class V | 5.72 – 7.15 | 17 – 28 |
|   | Class VI | 7.15 – 8.58 | 8 – 16 |
|   | Class VII | 8.58 – 10 | 1 – 7 |
|   | Class VIII | >10 | 0 |

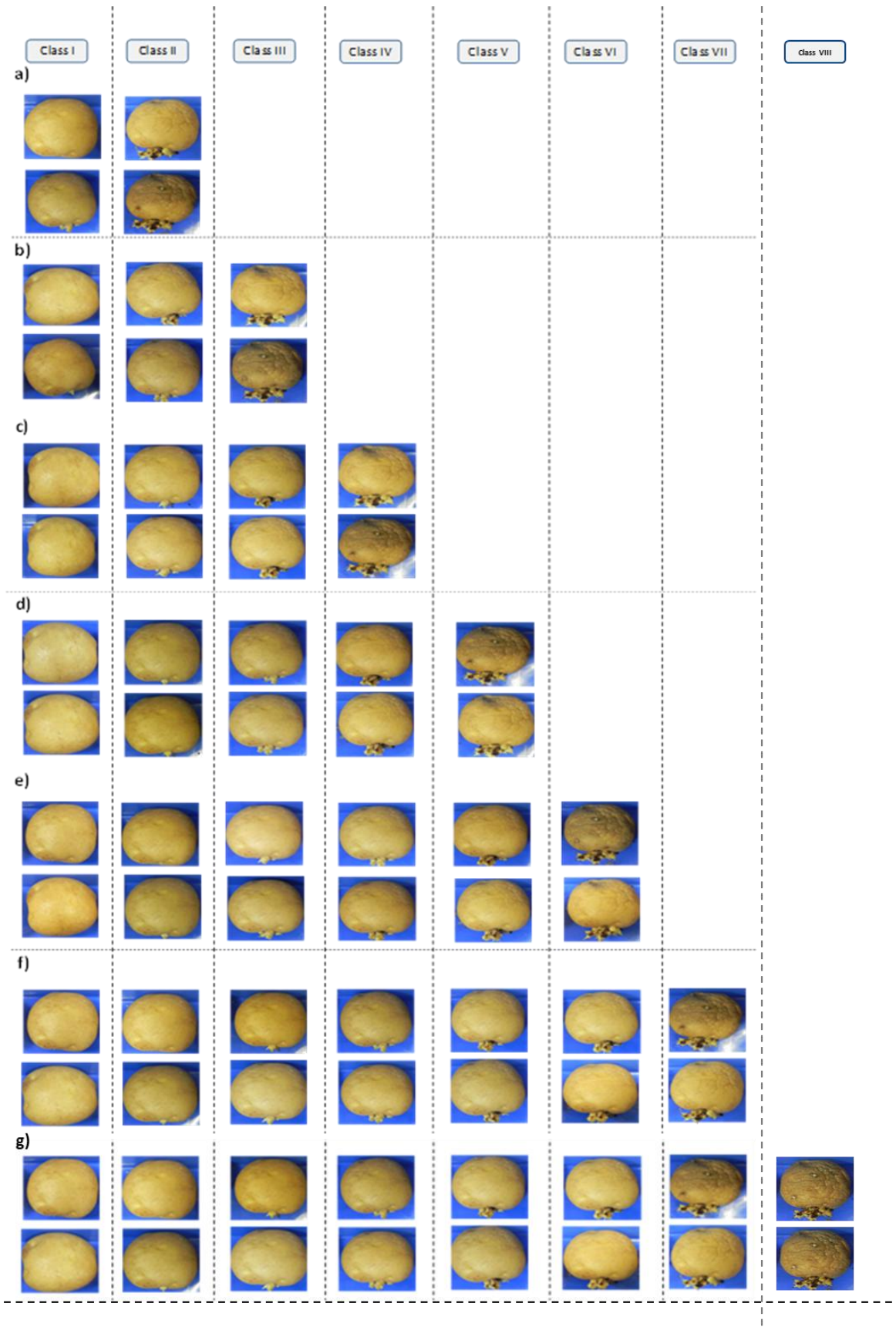

Figure 2: The sample images illustrating visual differences among potato classes in multiclass shelf-life prediction models (2–8 Classes)

**2.4 Model Architecture**

This study evaluates four state-of-the-art deep learning architectures: VGG, ResNet, DenseNet, and Vision Transformer (ViT) for multi-class classification of potato weight-loss categories. Each architecture represents a distinct family within computer vision, encompassing both conventional convolutional neural networks (CNNs) and modern transformer-based vision models. All models were initialized with pre-trained ImageNet weights (Russakovsky et al., 2015) to leverage transfer learning and subsequently fine-tuned for the task of multi-class weight loss prediction in potatoes. The structural details and modifications applied to each architecture are discussed in the following subsections.

**VGG**

VGG, introduced by the Visual Geometry Group at the University of Oxford, is one of the most well-established convolutional neural network architectures for image recognition (Simonyan & Zisserman, 2014). It follows a simple, yet effective design based on a deep stack of convolutional layers employing small 3×3 filters with consistent stride and padding. Max-pooling layers are inserted periodically to reduce spatial resolution and extract higher-level features.

Among its variants, VGG-16 and VGG-19 are the most widely used, containing 16 and 19 weight layers, respectively, encompassing both convolutional and fully connected layers. In this study, the VGG-16 model was chosen for its balanced trade-off between accuracy and computational cost. To tailor the model for the potato weight loss classification task, the original 1000-class output layer was replaced with two fully connected layers comprising 1024 neurons, followed by a Softmax layer that generates probabilities across the number of output classes ($n$), as shown in figure 3. The modified VGG-16 model contains approximately 14.89 million trainable parameters.

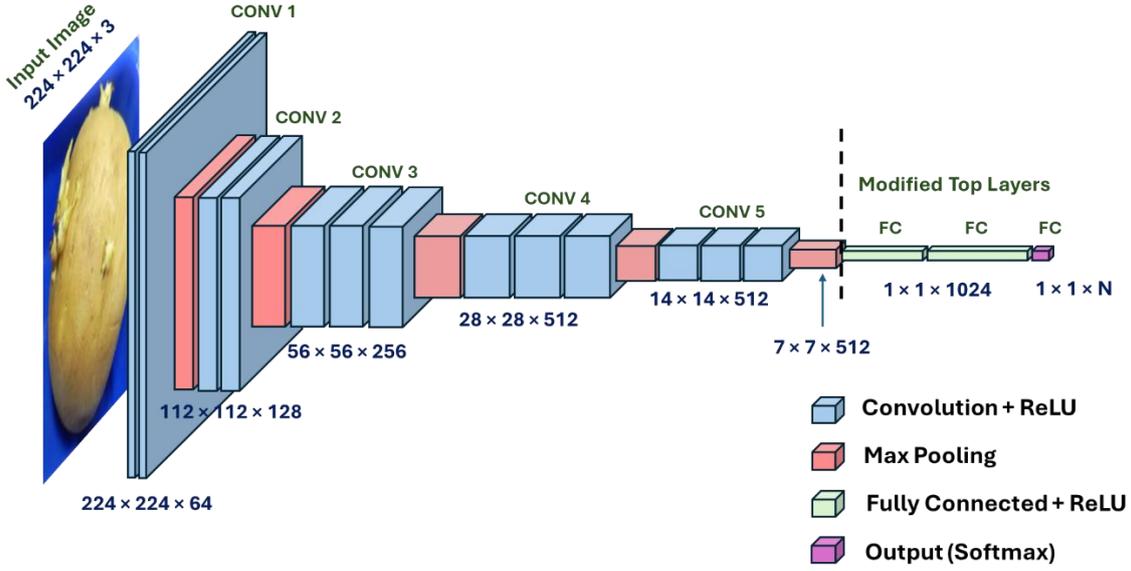

Figure 3: VGG-16 Architecture with modified top layers

### ResNet

Residual Network (ResNet) was introduced by He et al. (2016) to address the vanishing gradient problem that hampers the training of very deep CNNs. The key innovation in ResNet is the residual connection, or "shortcut," which allows the input of a layer to bypass one or more intermediate layers and be added directly to the output. This enables stable training of networks with hundreds of layers and improves feature reuse without increasing computational complexity.

Each residual block is mathematically defined as:

$$y = F(x, \{W_i\}) + x \qquad (1)$$

Where x is the input, $F(x, \{W_i\})$ represents the residual mapping to be learned, and y is the block output. The addition of x ensures identity mapping and eases gradient flow during backpropagation.

In this study, ResNet-50 was employed as a representative deep residual architecture. The model comprises an initial convolutional layer with $7 \times 7$ filters followed by max-pooling, and 4 stages of bottleneck residual blocks with 3, 4, 6 and 3 blocks, respectively, resulting in a total of 50 weight layers. The output of each stage is progressively down-sampled, while the number of feature channels increases, enabling the network to learn rich hierarchical representations. The top classification layers were modified to match the $n$-class output. The ResNet-50 model contains approximately 23.52 million trainable parameters. Figure 4 illustrates the modified ResNet-50 architecture used in this work, showing the convolutional layers, bottleneck residual blocks at each stage, down-sampling operations, and the final fully connected classification layer.

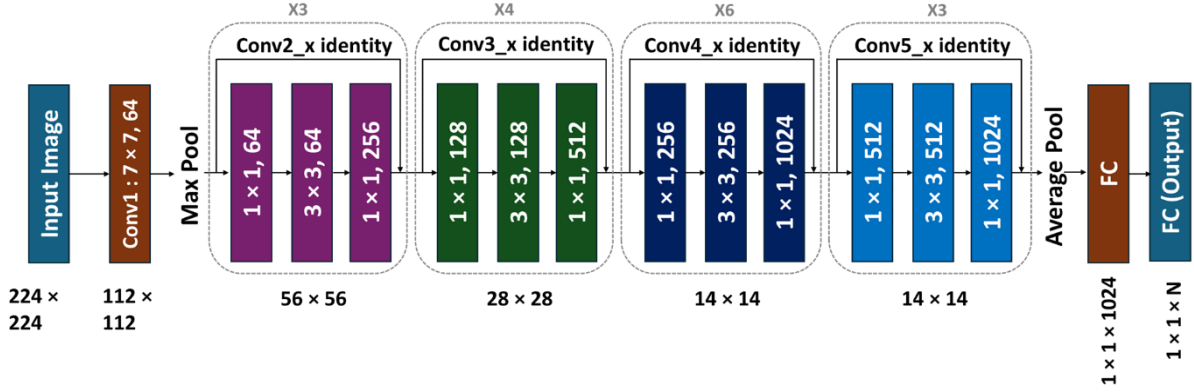

Figure 4: ResNet-50 Detailed Architecture with modified top layers

**DenseNet**

Dense Convolutional Network (DenseNet), proposed by Huang et al. (2017), introduces a dense connectivity pattern in which each layer receives feature maps from all preceding layers. Formally, the $l^{th}$ layer receives as input the concatenation of all feature maps from previous layers:

$$x_l = H_l([x_0, x_1, x_2, \ldots, x_{l-1}]) \tag{2}$$

Where $H_l(\ )$ represents a composite function of batch normalization, ReLU activation, and convolution operations. This dense connectivity enhances information flow, mitigates the vanishing gradient problem, and substantially reduces the number of parameters by promoting feature reuse.

The DenseNet-121 model used in this study comprises four dense blocks separated by transition layers that perform convolution and pooling operations. The final global average pooling layer was followed by fully connected layers (1024 neurons each) and an output Softmax layer for classification as shown in figure 5. The DenseNet-121 model contains approximately 6.96 million trainable parameters, nearly 17 times fewer than VGG-16, making it computationally efficient for large-scale classification.

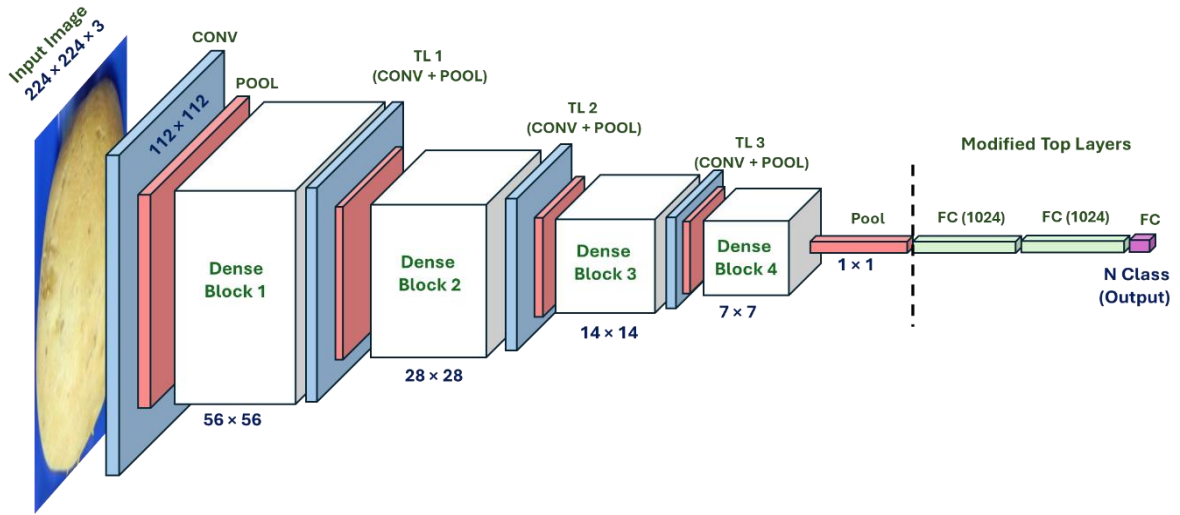

Figure 5: DenseNet-121 Architecture with modified top layers

**Vision Transformer**

Inspired by the remarkable performance of Transformers in natural language processing, the Vision Transformer (ViT) proposed by *Dosovitskiy et al.* (2020) adapts the self-attention mechanism for visual recognition tasks. Unlike convolution-based architectures, ViT processes images by splitting a 224 × 224 input into a set of fixed-size, non-overlapping patches (typically 16 × 16). Each patch is flattened and mapped to a vector representation through a linear projection, forming a sequence of patch embeddings. To retain spatial context, positional encodings are added to these embeddings. In addition, a learnable class token is prepended to the sequence, allowing the transformer encoder to aggregate global image information, as illustrated in figure 6.

The key component of ViT is the multi-head self-attention mechanism (Vaswani et al., 2017), which enables the model to capture long-range dependencies between different image regions. The attention operation is mathematically expressed as:

$$Attention\,(Q,K,V) = softmax\left(\frac{QK^T}{\sqrt{d_k}}\right)V \qquad (3)$$

Where Q, K and V denote the query, key and value matrices, and $d_k$ represents the dimension of the key vectors. The multi-head formulation performs multiple parallel attention computations (*h* heads), whose outputs are concatenated and linearly projected. Each transformer encoder block also incorporates residual connections (He et al., 2016) and layer normalization to stabilize the learning process

In this study, the ViT-B/16 configuration was utilized, comprising 12 transformer encoder layers, 12 attention heads ($h = 12$) and an embedding dimension of 768. The classification head was customized with two fully connected layers (each containing 1024 neurons) followed by a Softmax layer for 10-class prediction. The ViT-B/16 model contains approximately 86.6 million trainable parameters. The overall network structure used in this study is depicted in figure 6.

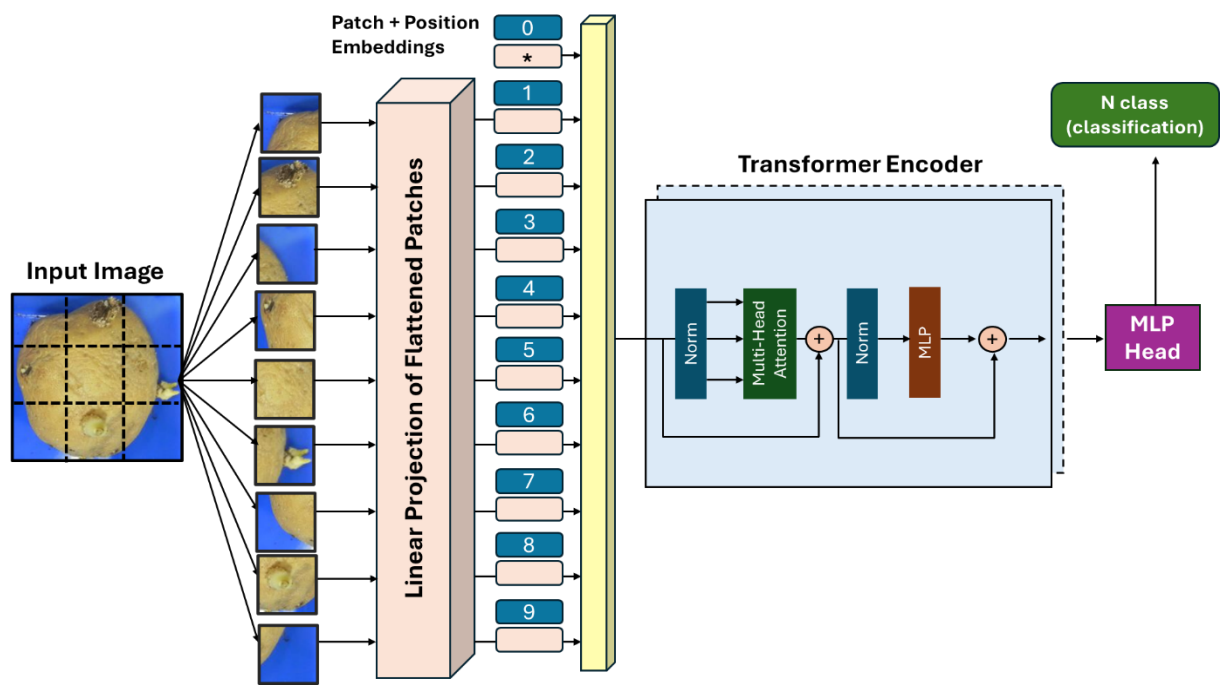

Figure 6: Vision Transformer (ViT) detailed architecture

## 2.5 Model Training

The training process employed a comprehensive strategy to ensure robust performance and generalization. All models were trained for up to 500 epochs with a batch size of 16 and an input image resolution of 224 × 224 pixels. The Adam optimizer was used with a learning rate of 0.001 for CNN-based architectures and 0.0001 for Vision Transformer (ViT). Cross-Entropy Loss with label smoothing ($\varepsilon = 0.1$) was adopted to mitigate overconfidence in predictions. A learning rate scheduler (ReduceLROnPlateau) was applied with a reduction factor of 0.5 and patience of 30 epochs, while early stopping was implemented with a patience of 100 epochs to prevent overfitting. Hyperparameter tuning was performed using GridSearchCV, and model evaluation utilized 5-fold stratified cross-validation to compute mean accuracy and standard deviation across folds. For each fold, a fresh model was initialized and trained, and performance metrics including accuracy, F1-score, precision, and recall were recorded. Data augmentation was applied using PyTorch's default transforms, including random resized cropping, horizontal and vertical flips, rotation (±20°), color jitter, and affine transformations, while avoiding heavy augmentations to maintain convergence stability. Pre-trained backbones such as VGG-16, ResNet-50, DenseNet-121 and ViT-B/16 were fine-tuned with a custom classifier adapted to the target number of classes. Final evaluation was conducted on the test subset of each fold, and confusion matrices were generated for detailed error analysis.

## 2.6 Model performance evaluation

The image-based models developed were validated using unseen test datasets. For all the models, test data was around 20% of total data. The model performance was assessed using confusion matrix which compares number of correct predictions with number of incorrect predictions for every class used in the model. The values of True Positive (TP), True Negative (TN), False Positive (FP) and False Negative (FN) were noted from the confusion matrix and subsequent estimation of statistical parameters such as precision, recall, F-score and classification accuracy were done using formulae (Equation (3) to (6)) as given below:

$$Precision = \frac{TP}{TP + FP} \tag{3}$$

$$Recall = \frac{TP}{TP + FN} \tag{4}$$

$$F1\ score = \frac{2 \times Precision \times Recall}{Precision + Recall} \tag{5}$$

$$Classification\ accuracy = \frac{Number\ of\ correctly\ classified\ potatoes}{Total\ number\ of\ samples} \tag{6}$$

The model performance for sprout and non-sprout classification was also evaluated using heat map. A heat map shows which parts of an input image were looked at by the model for assigning the class. The heat map was obtained for random samples chosen from test dataset for each of the two binary models. These maps were generated using Gradient-weighted Class Activation Mapping (Grad-CAM) algorithm, which uses the gradient of the predicted class score to obtain neuron importance weights. A weighted combination of these weights followed by a Rectified Linear Unit (ReLU) gives the class activation maps. When visualized as a heat map, these class activation maps provide us with an understanding of how the model assigns importance to various parts of the image when predicting the classification for the image. This can correlate with the locations of the specific type of damage on the potato that we are looking for, such as sprouts or fungus formations.

## 3. Results and Discussion

### 3.1 Top layer fine tuning

VGG-16, ResNet-50, DenseNet-121, and Vision Transformer (ViT-B/16) were employed for developing the classification model. Figure 7 illustrates the impact of different top-layer architectures on classification accuracy. The evaluation began with a configuration having no additional top layer, denoted as NoTop-4, followed by architectures with one top layer of 1024 neurons (1024-4), and then two and three top layers (1024-1024-4 and 1024-1024-1024-4). Configurations such as 1024-4 and 1024-1024-4 achieved the highest accuracy across most models, whereas the NoTop-4 setup caused a significant drop for VGG-16 (below 0.3), highlighting the importance of fine-tuning top layers for effective adaptation of features. DenseNet-121 and ViT-B/16 demonstrated strong performance across all configurations, indicating better generalization and robustness compared to VGG-16, and ResNet-50.

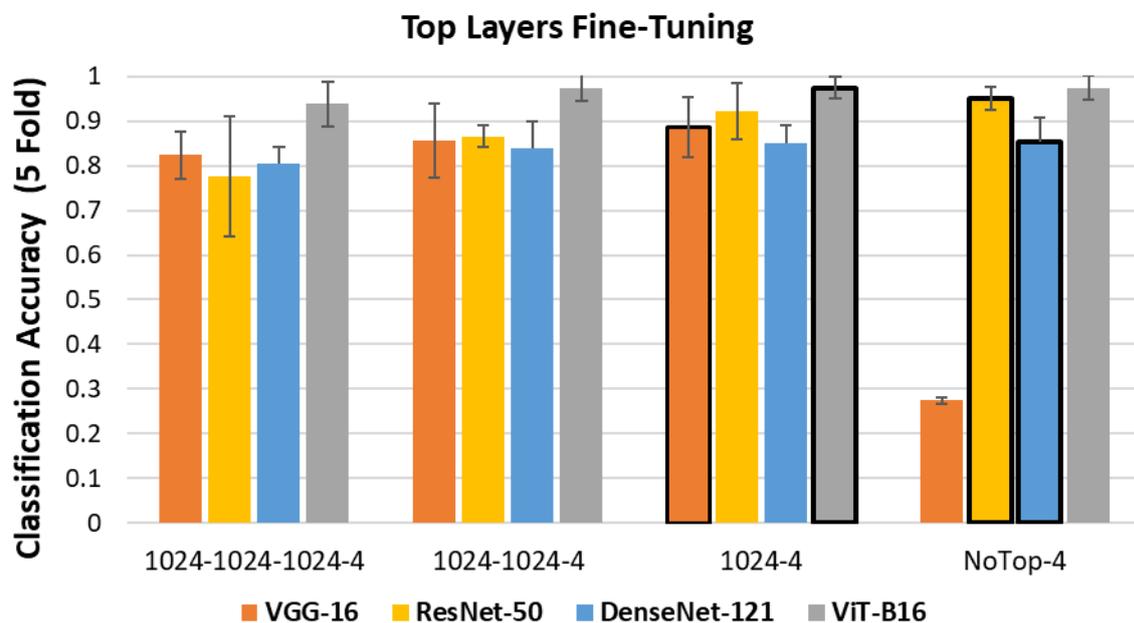

Figure 7: Effect of top-layer fine-tuning on classification accuracy

Figure 8 presents the model training and loss curves. Due to the application of learning rate decay, the fluctuations in loss progressively diminish as the models approach convergence. Early stopping with a patience of 100 epochs was employed, and all models trained for approximately 500 epochs, ensuring sufficient learning and reducing bias. Additionally, dropout with a rate of 0.5 and batch normalization were applied to the top layers to prevent overfitting and promote a more regularized model. It can also be observed that ViT converges significantly faster compared to the other models, indicating its efficiency in learning representations (Figure 8d).

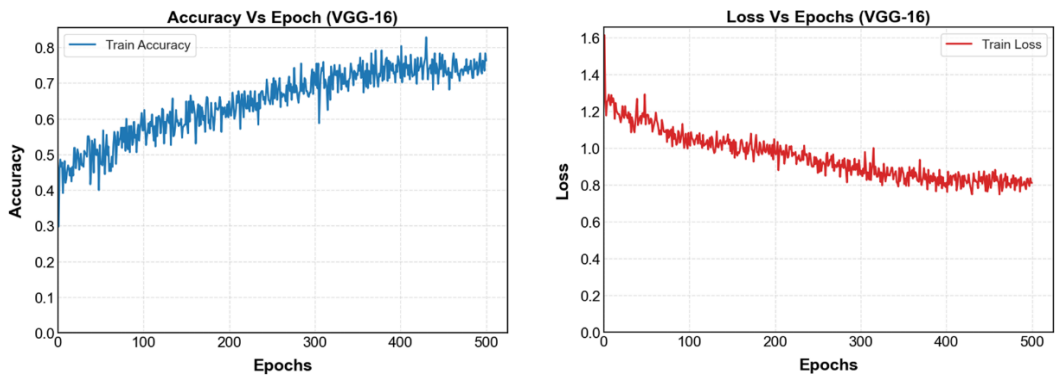

a) VGG-16

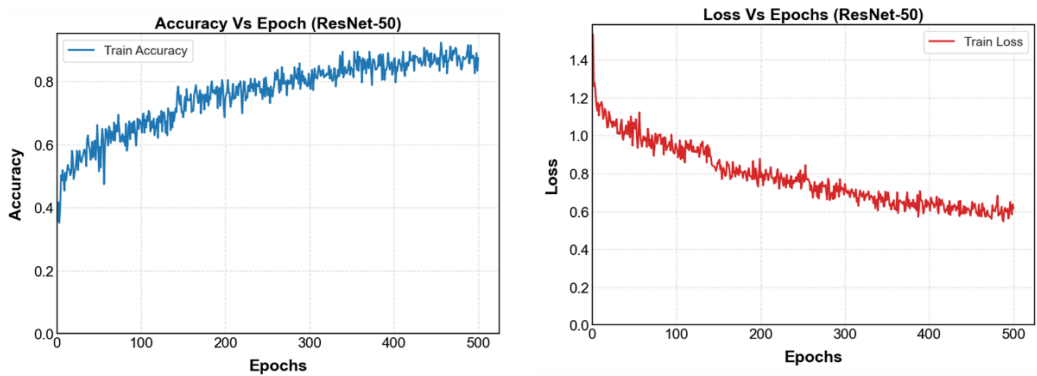

b) ResNet-50

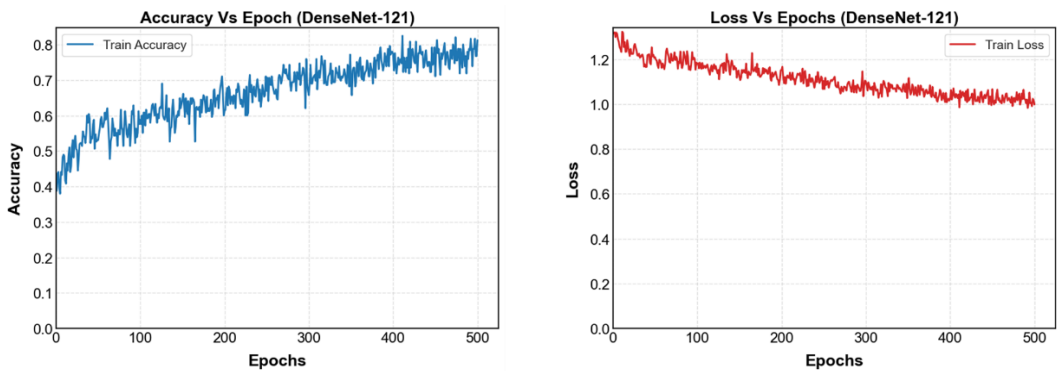

c) DenseNet-121

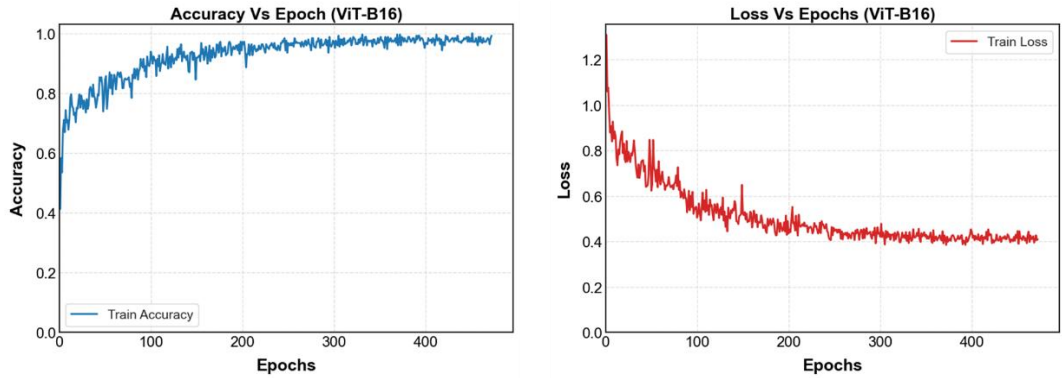

d) ViT-B/16

Figure 8: History plots a) VGG-16 b) ResNet-50 c) DenseNet-121 d) ViT-B/16

### 3.2 Potato sprout Classification

A sprout classification model was developed to detect sprouting in potatoes using annotated images labeled as either "Sprouts" or "Non-sprouts." The dataset images of potatoes capturing a range of sprouting stages from no visible sprouts to small, localized sprouts and extensive sprouting across the potato surface. The dataset comprised 255 training images and 51 test images. The model was evaluated on a separate test set, and its performance was assessed using a confusion matrix. The training process was monitored using loss and accuracy curves across epochs.

The sprout classification task was formulated as a binary classification problem using four different architectures to evaluate sprout detection accuracy. Table 2 summarizes the performance metrics such as accuracy, precision, recall, and F1-score for all four models, reported as the mean across five folds along with their standard deviations. Furthermore, figure 9 illustrates the overall classification accuracy for sprout/non-sprout detection in the form of a histogram with error bars, providing a visual comparison of model performance. All models achieved near-perfect accuracy, with DenseNet-121, ResNet-50, ViT-B/16 reaching 98% and VGG-16 slightly lower at 92%. These results confirm that CNN and ViT based deep learning models can reliably identify sprouting in potatoes regardless of sprout size or distribution.

The superior performance of DenseNet-121 and ResNet-50 can be attributed to their ability to capture fine-grained local features through dense connectivity and residual learning. ViT's competitive performance highlights the strength of transformer-based architectures in modeling global dependencies, which is advantageous for detecting distributed sprouting patterns across the potato surface. The lower performance of VGG-16 under limited fine-tuning suggests that older CNN architectures may struggle with complex visual variations compared to modern deep networks. Fine-tuning strategies significantly influenced performance, with partial layer adaptation yielding the best results. DenseNet-121 and ResNet-50 remain highly effective for localized feature detection, while ViT demonstrates strong potential for agricultural image analysis, especially when global context matters.

**Table 2:** Comparative analysis of model architectures for Sprout classification

| Evaluation Matrix | VGG-16 | | ResNet-50 | | DenseNet-121 | | ViT-B/16 | |
|---|---|---|---|---|---|---|---|---|
| | Mean | STD | Mean | STD | Mean | STD | Mean | STD |
| Accuracy | 0.9181 | 0.0544 | 0.9771 | 0.0187 | 0.9803 | 0.0137 | 0.9804 | 0.0179 |
| F1 Score | 0.9118 | 0.0624 | 0.9773 | 0.0182 | 0.9804 | 0.0137 | 0.9803 | 0.0181 |
| Precision | 0.9291 | 0.0432 | 0.9793 | 0.0159 | 0.9816 | 0.0126 | 0.9806 | 0.0181 |
| Recall | 0.91813 | 0.0545 | 0.9771 | 0.0018 | 0.9803 | 0.0137 | 0.9804 | 0.0179 |

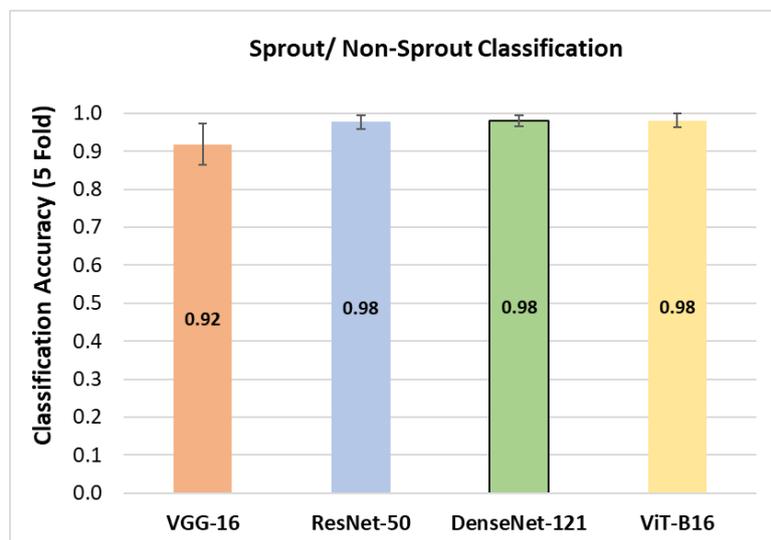

Figure 9: Comparison of overall classification accuracy for sprout/non-sprout detection

The confusion matrix demonstrates near-perfect separation between classes: the model correctly identified 18/18 non-sprouts and 42/43 sprouts (TN=18, FP=0, FN=1, TP=42), yielding accuracy 98.36% and balanced accuracy 98.84% (Figure 10). The absence of false positives (FP=0) indicates perfect specificity, while a single missed sprout constrains sprout recall to 97.67% (precision 100.00%, F1 98.82%). For the non-sprout class, precision 94.74% and recall 100.00% (F1 97.30%) show minimal label leakage. The MCC of 0.962 confirms strong agreement despite the fold's class imbalance (43 sprouts vs. 18 non-sprouts). Practically, the error profile is dominated by one false negative sprout, suggesting a conservative decision boundary; if the application prioritizes recovering every sprout, a modest threshold shift or cost-sensitive training could trade a small increase in false positives for eliminating these rare misses.

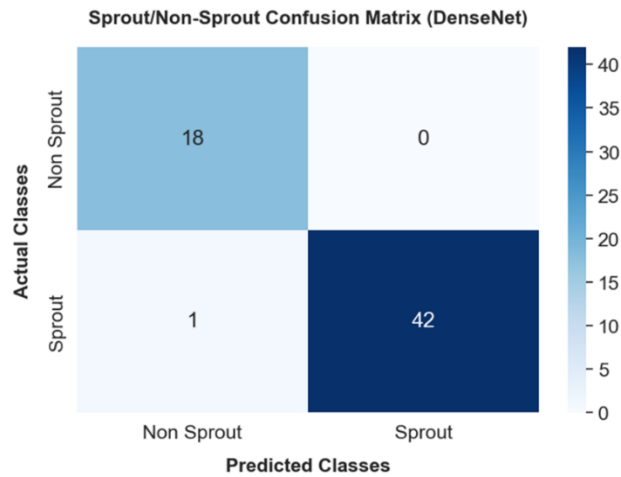

**Figure 10:** Confusion matrix for Sprout/Non-Sprout Classification using DenseNet-121

**Heatmap Visualization of CNN Based Architecture using Grad-CAM (Gradient Class Activation Maps):**

The Grad-CAM heat maps highlight the image regions that influenced the classification decisions of different CNN architectures for sprouted potatoes. Figure 10 presents class activation maps for four sample images across all trained models.

In figure 11 (a), all three CNNs correctly emphasize the sprout region before classifying the image as "sprouted." However, ResNet occasionally highlights areas unrelated to sprouts, while VGG produces a more dispersed attention map. DenseNet demonstrates the most precise localization in this case. In figure 11 (b), similar patterns are observed, with DenseNet focusing sharply on the large sprout, whereas VGG and ResNet show broader attention. In figure 11 (c), VGG misclassifies the image as "non-sprout," resulting in a nearly flat heat map. The small sprout size likely caused this error. DenseNet, in contrast, accurately highlights the sprouted pixels. In figure 11 (d), VGG highlights three regions, two corresponding to sprouts and one to surface irregularities, while ResNet misses one sprout. DenseNet again provides the most precise activation.

Overall, DenseNet consistently delivers better localization of sprout features compared to VGG and ResNet. VGG tends to distribute attention across the potato surface, indicating less precise feature focus. These observations confirm that deeper architectures like ResNet and DenseNet offer superior interpretability and feature sensitivity, which is critical for reliable classification in agricultural quality assessment.

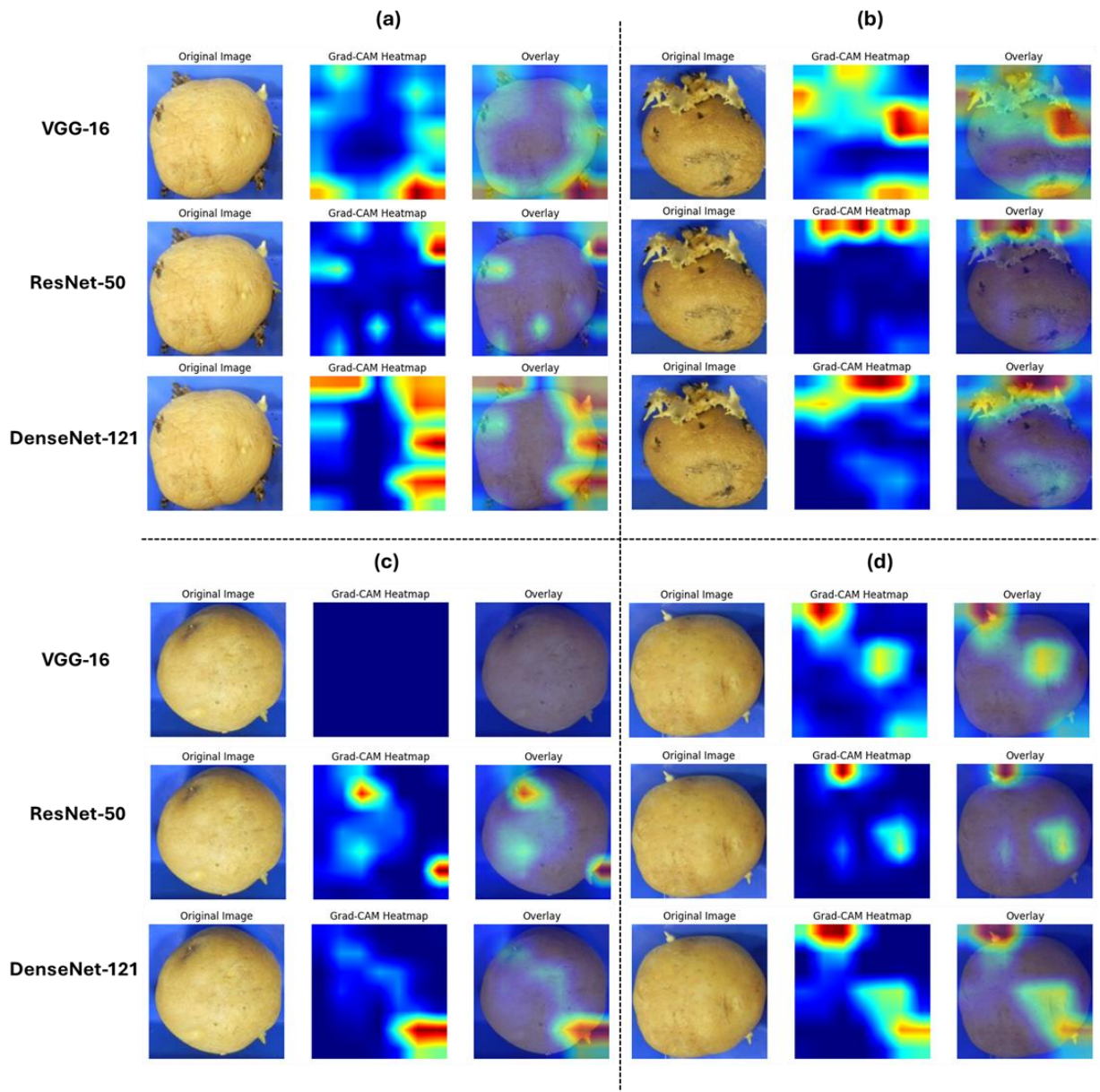

Figure 11: Heatmap Visualization

## 3.3 Shelf-life prediction (Multi-class classification model)

For each potato, images were captured daily over a 200-day storage period, and each image was annotated with its corresponding weight loss percentage and remaining shelf-life. The trained models predict the class (i.e., the range) in which a given potato's current weight loss and remaining shelf-life fall. The dataset comprised images from individual potatoes. 5-fold cross validation was used, where data from 255 images were used for training, and 51 images were used for testing.

Figure 12 illustrates the impact of increasing the number of classes on model accuracy. Accuracy decreased almost linearly from ~99% for 2-class models to ~82% for 7-class models. This decline is primarily due to two factors:

(a) Reduced visual distinction between adjacent classes as the number of classes increases, making it harder for the model to learn discriminative features.

(b) Fewer samples per class with more classes, since the overall dataset size remains constant.

Among the architectures, ViT-B/16 achieved the highest accuracy across most class configurations, followed by DenseNet-121 and ResNet-50, VGG-16 consistently lagged behind, especially for higher class counts. These results suggest that transformer-based models are better suited for capturing global contextual features, while CNN-based models remain competitive for localized visual cues (For example Table 3 (5 class classification)).

Multi-class shelf-life prediction is feasible with high accuracy for coarse class divisions (2–5 classes), but performance declines with finer granularity. For practical applications, models with fewer classes may offer a better trade-off between accuracy and usability.

It is evident that accuracy decreases almost linearly from 99.01% to 82.68% as the number of classes increases from 2 to 7 (Figure 12). This is primarily due to two reasons: a) The visual differences between adjacent classes become less distinct as the number of classes increases, making it harder for the model to learn meaningful patterns b) The number of training samples per class decreases with more classes, since the overall dataset size remains nearly constant across all models.

**Table 3:** Comparative analysis of model architectures for 5 class classification

| | 5 Class Classification | | | | | | | |
| --- | --- | --- | --- | --- | --- | --- | --- | --- |
| | DenseNet-121 | | VGG-16 | | ResNet-50 | | ViT-B/16 | |
| **Evaluation Matrix** | **Mean** | **STD** | **Mean** | **STD** | **Mean** | **STD** | **Mean** | **STD** |
| **Accuracy** | 0.8918 | 0.0298 | 0.7213 | 0.0418 | 0.8131 | 0.0377 | 0.8984 | 0.0618 |
| **F1 Score** | 0.8908 | 0.0308 | 0.6911 | 0.0522 | 0.8043 | 0.0462 | 0.8975 | 0.0627 |
| **Precision** | 0.9015 | 0.0236 | 0.7074 | 0.0811 | 0.8545 | 0.0287 | 0.9045 | 0.0618 |
| **Recall** | 0.8918 | 0.0298 | 0.7213 | 0.0418 | 0.8131 | 0.0377 | 0.8984 | 0.0618 |

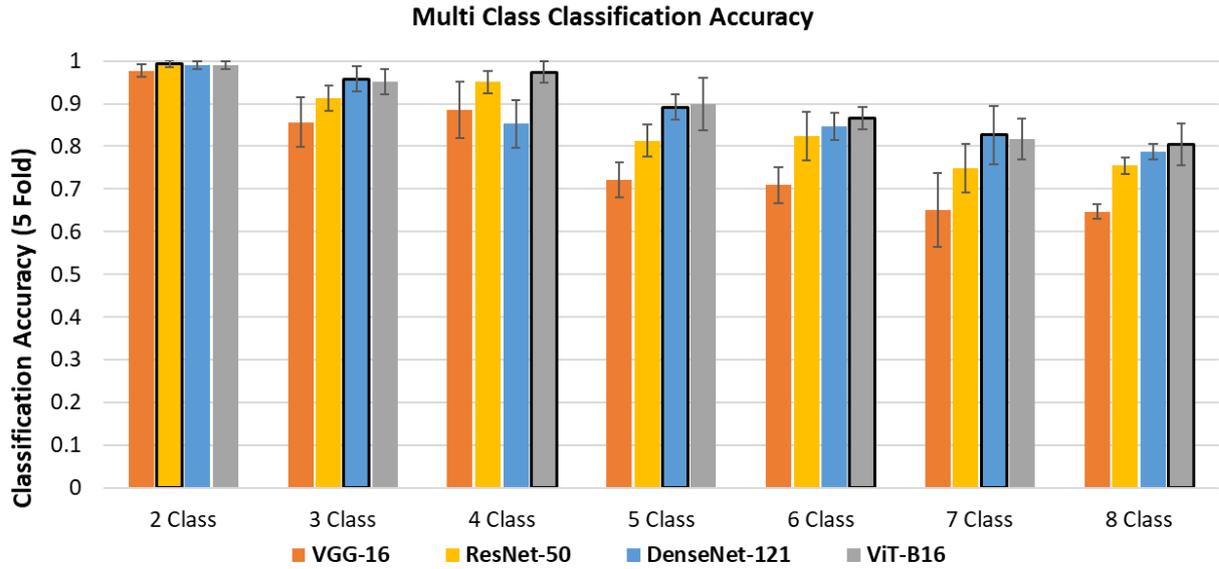

Figure 12: Comparison of overall classification accuracy for multi-class classification

The confusion matrices indicate that the proposed model demonstrates strong classification performance across class intervals, with predictions concentrated along the diagonal and minimal off-diagonal errors (Figure 13). Most instances are correctly classified into their respective percentage ranges, reflecting high precision and recall for each category. Misclassifications, where present, are limited to adjacent classes, suggesting that the model effectively captures the underlying patterns but faces slight challenges in distinguishing closely related intervals. Overall, the results confirm the robustness and reliability of the approach for accurate percentage-based categorization.

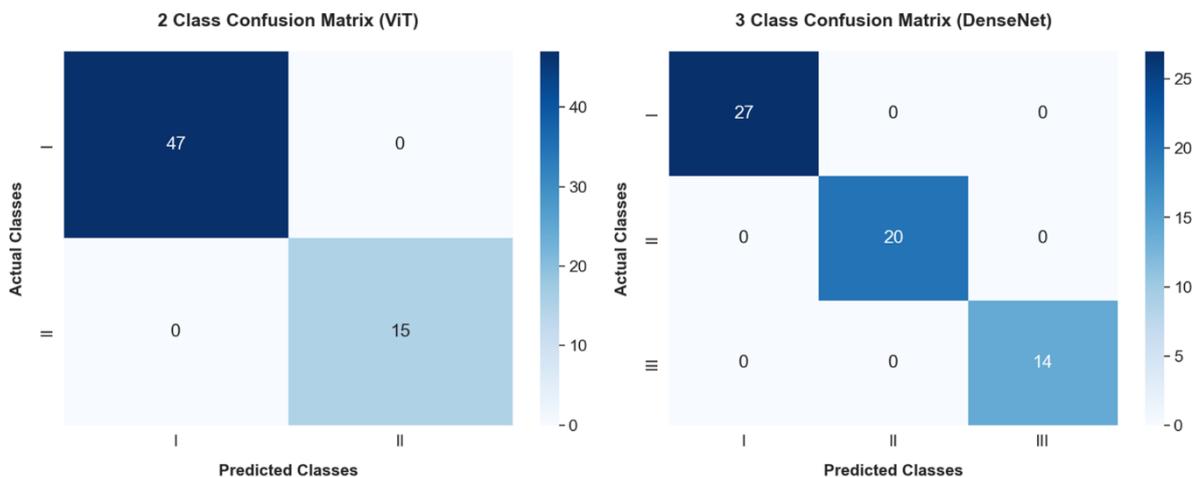

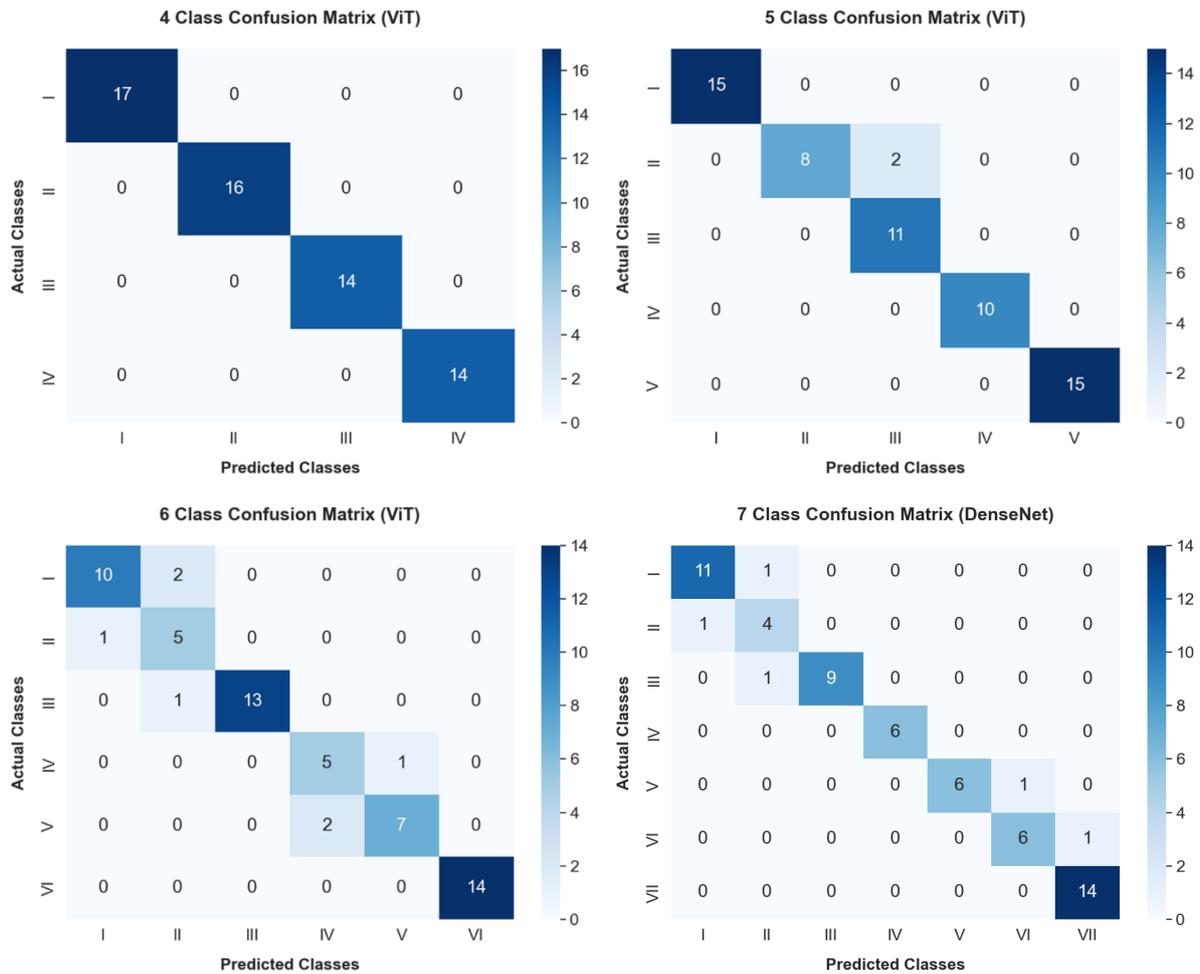

Figure 13: Confusion matrix for best performing model for 2-7 class classification

There is numerous potato varieties cultivated globally, each differing in characteristics such as nutritional composition, dry matter content, shape, size, color, and genetic traits. These differences influence how each variety behaves during storage. Several studies have reported significant variation in parameters like weight loss, dormancy period, and sprouting across different cultivars (Gupta et al., 2015; Pande et al., 2007). For instance, Gupta et al. (2015) observed dormancy periods ranging from 45 to 75 days and weight loss between 7.5% and 17.5% after 75 days of storage across 44 potato varieties.

Such variability implies that the true shelf-life of a potato may differ from the shelf-life predicted by models trained on a specific variety. Therefore, the models developed in this study should be applied to other potato varieties with caution. For broader applicability, a more generalized model trained on a diverse dataset comprising multiple potato varieties is recommended.

## 3.4 Computational Complexity and Efficiency Analysis

Table 4 summarizes the architectural complexity and computational performance of the evaluated models. DenseNet-121 is the most parameter-efficient network with 6.96M parameters and 2.9 GFLOPs, though its inference time (0.01219 s/image) is slightly higher than ResNet-50 and VGG-16. ResNet-50, despite having 23.52M parameters, achieves the fastest training time (13.16 min) and competitive inference latency (0.00486 s/image), indicating a favorable trade-off between depth and efficiency. VGG-16, with 14.89M parameters, incurs significantly higher computational cost (15.4 GFLOPs) and longer training time (16.95 min), reflecting its less optimized architecture. ViT-B/16 exhibits the highest resource demand (86.6M parameters, 17.61 GFLOPs) and longest training time (19.45 min), consistent with transformer-based models' complexity. These results highlight ResNet-50 and DenseNet-121 as practical choices for real-time potato sprout classification, balancing accuracy, interpretability, and computational efficiency (Table 4).

**Table 4:** comparison of model architecture based on computational complexity

| Model | Parameters | GFLOPs | Avg. Training Time (min) (1-fold) | Avg. Inference Time (sec/image) |
|---|---|---|---|---|
| DenseNet-121 | 6.96 M | 2.9 GMac | 15.72 | 0.01219 |
| ResNet-50 | 23.52 M | 4.13 GMac | 13.16 | 0.00486 |
| VGG-16 | 14.89 M | 15.4 GMac | 16.95 | 0.00191 |
| ViT-B/16 | 86.6 M | 17.61 GMac | 19.45 | 0.00422 |

## 3.5 Human Accuracy Evaluation

To establish a benchmark for model performance and assess the potential for improvement, human accuracy study was conducted on the same dataset used for multi-class classification. Five domain experts independently annotated images for the 4-class classification scenario. The aggregated human performance metrics were as follows: Accuracy = 0.9020, Precision = 0.9139, Recall = 0.9020, and F1 Score = 0.9079. These results indicate that while our models: DenseNet-121, VGG-16, ResNet-50, and ViT-B/16, achieved high accuracy in binary classification (up to 99%), their performance declined to approximately 82% for 7-class classification. The human benchmark underscores the inherent complexity of the task and provides a realistic upper bound for achievable accuracy on this dataset. This comparison is critical for interpreting model performance and guiding future improvements.

## 3.6 Practical Implications

The high accuracy achieved by image-based deep learning models for automated sprout detection and multi-class shelf-life prediction in potatoes demonstrates their strong potential for transforming storage and supply chain management. By integrating these models into sorting systems, sprouted potatoes can be identified early, reducing manual inspection efforts and enabling timely decisions for sale, processing, or disposal ultimately minimizing economic losses and food waste. Similarly, shelf-life prediction models with coarse class divisions (2–5 classes) allow reliable categorization of potatoes into early, mid, or late storage stages, supporting dynamic inventory management and differential pricing strategies that optimize value for consumers while reducing spoilage costs for suppliers. Automated image-based classification enhances efficiency and scalability in large facilities, replacing labor-intensive manual checks. However, the decline in accuracy for finer class divisions (6–8 classes) highlights the challenge of predicting exact intervals due to subtle visual differences and limited data per class. Therefore, practical deployment should focus on fewer, broader classes for robust performance, while future research should aim to develop generalized models trained on diverse potato varieties and storage conditions to improve adaptability across supply chains.

## 4. Conclusions

This study demonstrates the effectiveness of image-based models in screening and predicting the shelf-life of potatoes within the supply chain. Two distinct models were developed: a sprout detection model and a shelf-life prediction model. The sprout detection model classified potatoes based on the presence or absence of sprouts with an accuracy exceeding 98%. The shelf-life prediction model categorized potatoes into 2, 3, 4, 5, 6, or 7 classes, achieving average accuracies of 99.34%, 95.76%, 97.38%, 89.18%, 86.60%, and 82.68%, respectively. These results indicate that dividing post-harvest life into 2, 3, or 5 classes provides reliable accuracy greater than 89%. The study also revealed that model performance depends on the variability among potatoes used for training and the size of the dataset, emphasizing the need for large and diverse datasets to improve generalization and prediction accuracy.

Future research should focus on integrating visual indicators of weight loss with physicochemical properties such as vitamin content, pH, dry matter, and phenolic compounds to develop comprehensive models capable of predicting multiple quality attributes from images. Additionally, the approach can be extended to other root and tuber crops, including sweet potatoes, carrots, radishes, and beets, which undergo similar quality degradation during storage and transportation. Developing generic models for the entire class of roots and tubers will enable prediction across diverse species by accounting for physiological processes such as respiration, transpiration, dormancy break, and mechanical damage under varying environmental conditions.

**CRediT authorship contribution statement**

**Shrikant Kapse:** Conceptualization, Data curation, Methodology, Formal analysis, Investigation, Validation, Writing – original draft, Writing – review & editing. **Priyankkumar Dhrangdhariya:** Formal analysis, Methodology, Investigation, Writing – original draft, Writing – review & editing. **Priya Kedia:** Data curation, Writing – original draft, Writing – review & editing. **Manasi Patwardhan:** Data curation, Formal analysis, Methodology. **Shankar Kausley:** Investigation, Methodology, Writing – original draft, Writing – review & editing. **Beena Rai:** Conceptualization, Methodology, Supervision. **Shirish Karande:** Conceptualization, Methodology.

**Declaration of Competing Interest**

The authors declare that they have no known competing financial interests or personal relationships that could have appeared to influence the work reported in this paper.

**Acknowledgements**

The authors gratefully acknowledge the support provided by Dr. Sachin Lodha, Head, TCS Research and Dr. Harrick Vin, CTO, Tata Consultancy Services during this project. The authors also express gratitude towards lab assistants Mr. Sudam Konelu, Mr. Rupesh Shinde and Mr. Yaseen Shaikh for helping with data collection from in-house experiments.

# References

Alhammad, S. M., Khafaga, D. S., El-Hady, W. M., Samy, F. M., & Hosny, K. M. (2025). Deep learning and explainable AI for classification of potato leaf diseases. *Frontiers in Artificial Intelligence, 7,* 1449329. https://doi.org/10.3389/frai.2024.1449329

Azizi, A., Abbaspour-Gilandeh, Y., Nooshyar, M., & Afkari-Sayah, A. (2016). Identifying potato varieties using machine vision and artificial neural networks. *International Journal of Food Properties, 19*(3), 618–635. https://doi.org/10.1080/10942912.2015.1038834

Dacal-Nieto, A., Vázquez-Fernández, E., Formella, A., Martin, F., Torres-Guijarro, S., & González-Jorge, H. (2009, November). A genetic algorithm approach for feature selection in potatoes classification by computer vision. In *2009 35th Annual Conference of IEEE Industrial Electronics* (pp. 1955–1960). IEEE. https://doi.org/10.1109/IECON.2009.5414871

Danielak, M., Przybył, K., & Koszela, K. (2023). The need for machines for the nondestructive quality assessment of potatoes with the use of artificial intelligence methods and imaging techniques. *Sensors, 23*(4), 1787. https://doi.org/10.3390/s23041787

Dhrangdhariya, P., Saini, P., Maiti, S., & Rai, B. (2025). Multi-class classification of paint/coating defects using transfer learning. *Engineering Applications of Artificial Intelligence, 156,* 111320. https://doi.org/10.1016/j.engappai.2025.111320

Dosovitskiy, A. (2020). An image is worth 16x16 words: Transformers for image recognition at scale. *arXiv preprint arXiv:2010.11929.* https://arxiv.org/abs/2010.11929

Ebrahimi, E., Mollazade, K., & Arefi, A. (2012). An expert system for classification of potato tubers using image processing and artificial neural networks. *International Journal of Food Engineering, 8*(4). https://doi.org/10.1515/1556-3758.2656

Gülmez, B. (2025). A comprehensive review of convolutional neural networks based disease detection strategies in potato agriculture. *Potato Research, 68*(2), 1295–1329. https://doi.org/10.1007/s11540-024-09786-1

Gupta, V. K., Luthra, S. K., & Singh, B. P. (2015). Storage behaviour and cooking quality of Indian potato varieties. *Journal of Food Science and Technology, 52*(8), 4863–4873. https://doi.org/10.1007/s13197-014-1608-z

He, K., Zhang, X., Ren, S., & Sun, J. (2016). Deep residual learning for image recognition. In *Proceedings of the IEEE Conference on Computer Vision and Pattern Recognition* (pp. 770–778). https://doi.org/10.1109/CVPR.2016.90

Huang, G., Liu, Z., Van Der Maaten, L., & Weinberger, K. Q. (2017). Densely connected convolutional networks. In *Proceedings of the IEEE Conference on Computer Vision and Pattern Recognition* (pp. 4700–4708). https://doi.org/10.1109/CVPR.2017.243

Ibrahim, A., El-Bialee, N., Saad, M., & Romano, E. (2020). Non-destructive quality inspection of potato tubers using automated vision system. *International Journal on Advanced Science Engineering Information Technology, 10*(6), 2419–2428.

Kamilaris, A., & Prenafeta-Boldú, F. X. (2018). Deep learning in agriculture: A survey. *Computers and Electronics in Agriculture, 147,* 70–90. https://doi.org/10.1016/j.compag.2018.02.016

Le, T. T., & Lin, C. Y. (2019). Deep learning for noninvasive classification of clustered horticultural crops–A case for banana fruit tiers. *Postharvest Biology and Technology, 156,* 110922. https://doi.org/10.1016/j.postharvbio.2019.05.023

Nasiri, A., Taheri-Garavand, A., & Zhang, Y. D. (2019). Image-based deep learning automated sorting of date fruit. *Postharvest Biology and Technology, 153,* 133–141. https://doi.org/10.1016/j.postharvbio.2019.04.003

Pande, P. C., Singh, S. V., Pandey, S. K., & Singh, B. (2007). Dormancy, sprouting behaviour and weight loss in Indian potato (Solanum tuberosum) varieties. *Indian Journal of Agricultural Sciences, 77*(11).

Pinhero, R. G., Coffin, R., & Yada, R. Y. (2009). Post-harvest storage of potatoes. In *Advances in Potato Chemistry and Technology* (pp. 339–370). Academic Press. https://doi.org/10.1016/B978-0-12-374349-7.00012-X

Piedad, E. J., Larada, J. I., Pojas, G. J., & Ferrer, L. V. V. (2018). Postharvest classification of banana (Musa acuminata) using tier-based machine learning. *Postharvest Biology and Technology, 145,* 93–100. https://doi.org/10.1016/j.postharvbio.2018.06.004

Russakovsky, O., Deng, J., Su, H., Krause, J., Satheesh, S., Ma, S., ... & Fei-Fei, L. (2015). ImageNet large scale visual recognition challenge. *International Journal of Computer Vision, 115*(3), 211–252. https://doi.org/10.1007/s11263-015-0816-y

Simonyan, K., & Zisserman, A. (2014). Very deep convolutional networks for large-scale image recognition. *arXiv preprint arXiv:1409.1556.* https://arxiv.org/abs/1409.1556

Si, Y., Sankaran, S., Knowles, N. R., & Pavek, M. J. (2017). Potato tuber length-width ratio assessment using image analysis. *American Journal of Potato Research, 94*(1), 88–93. https://doi.org/10.1007/s12230-016-9545-1

Sonnewald, S., & Sonnewald, U. (2014). Regulation of potato tuber sprouting. *Planta, 239*(1), 27–38. https://doi.org/10.1007/s00425-013-1968-z

Su, Q., Kondo, N., Al Riza, D. F., & Habaragamuwa, H. (2020). Potato quality grading based on depth imaging and convolutional neural network. *Journal of Food Quality, 2020,* 8815896. https://doi.org/10.1155/2020/8815896

Su, W. H., & Xue, H. (2021). Imaging spectroscopy and machine learning for intelligent determination of potato and sweet potato quality. *Foods, 10*(9), 2146. https://doi.org/10.3390/foods10092146

Surya Prabha, D., & Satheesh Kumar, J. (2015). Assessment of banana fruit maturity by image processing technique. *Journal of Food Science and Technology, 52*(3), 1316–1327. https://doi.org/10.1007/s13197-013-1188-3

Vaswani, A., Shazeer, N., Parmar, N., Uszkoreit, J., Jones, L., Gomez, A. N., ... & Polosukhin, I. (2017). Attention is all you need. *Advances in Neural Information Processing Systems, 30,* 5998–6008.

Wang, Y., Deng, Y., Zheng, Y., Chattopadhyay, P., & Wang, L. (2025). Vision transformers for image classification: A comparative survey. *Technologies, 13*(1), 32. https://doi.org/10.3390/technologies13010032

Wei, Q., Zheng, Y., Chen, Z., Huang, Y., Chen, C., Wei, Z., ... & Chen, F. (2024). Nondestructive perception of potato quality in actual online production based on cross-modal technology. *International Journal of Agricultural and Biological Engineering, 16*(6), 280–290. DOI: https://doi.org/10.25165/j.ijabe.20231606.8076

Wu, H., Zhu, R., Wang, H., Wang, X., Huang, J., & Liu, S. (2025). Flaw-YOLOv5s: A lightweight potato surface defect detection algorithm based on multi-scale feature fusion. *Agronomy, 15*(4), 875. https://sdoi.org/10.3390/agronomy15040875